\documentclass[sigconf]{acmart}
\usepackage{multirow}
\usepackage{booktabs} \usepackage{amsmath}
\usepackage{breakurl}
\usepackage{subcaption}
\usepackage{enumitem,amssymb}
\usepackage{algorithm}
\usepackage{algorithmic}
\usepackage{tcolorbox} \newcommand{\ciao}[1]{{\setlength\fboxrule{0pt}\fbox{\tcbox[colframe=black,colback=white,shrink tight,boxrule=0.5pt,extrude by=1mm]{\small #1}}}}

\renewcommand\footnotetextcopyrightpermission[1]{}

\renewcommand{\vec}[1]{\mathbf{#1}} \newcommand{\vecg}[1]{\boldsymbol{#1}}  
\newcommand{\red}{\color{red}}

\DeclareMathOperator*{\argmin}{argmin}
\newcommand{\sect}{\textsection}

\acmConference[arXiv preprint] {preprint}{2019}{arXiv}

\begin{document}

\title[MTD for Embedded Deep Visual Sensing against Adversarial Examples]{Moving Target Defense for Embedded Deep Visual Sensing against Adversarial Examples}


\author{Qun Song}
\affiliation{Energy Research Institute, Interdisciplinary Graduate School}
\affiliation{School of Computer Science and Engineering\\Nanyang Technological University, Singapore}
\email{song0167@ntu.edu.sg}

\author{Zhenyu Yan}
\affiliation{School of Computer Science and Engineering, Nanyang Technological University, Singapore}
\email{zyan006@ntu.edu.sg}

\author{Rui Tan}
\affiliation{School of Computer Science and Engineering, Nanyang Technological University, Singapore}
\email{tanrui@ntu.edu.sg}

\begin{abstract}
  Deep learning based visual sensing has achieved attractive accuracy but is shown vulnerable to adversarial example attacks. Specifically, once the attackers obtain the deep model, they can construct adversarial examples to mislead the model to yield wrong classification results. Deployable adversarial examples such as small stickers pasted on the road signs and lanes have been shown effective in misleading advanced driver-assistance systems. Many existing countermeasures against adversarial examples build their security on the attackers' ignorance of the defense mechanisms. Thus, they fall short of following Kerckhoffs's principle and can be subverted once the attackers know the details of the defense. This paper applies the strategy of {\em moving target defense} (MTD) to generate multiple new deep models after system deployment, that will collaboratively detect and thwart adversarial examples. Our MTD design is based on the adversarial examples' minor transferability to models differing from the one (e.g., the factory-designed model) used for attack construction. The post-deployment quasi-secret deep models significantly increase the bar for the attackers to construct effective adversarial examples. We also apply the technique of serial data fusion with early stopping to reduce the inference time by a factor of up to 5 while maintaining the sensing and defense performance. Extensive evaluation based on three datasets including a road sign image database and a GPU-equipped Jetson embedded computing board shows the effectiveness of our approach.
\end{abstract}

\maketitle
\section{Introduction}

To implement autonomous systems operating in complex environments (e.g., the long envisaged self-driving cars), the accurate and resilient perception of the environment is often the most challenging step in the closed loop of sensing, decision, and actuation. The recent advances of deep learning \cite{lecun2015deep,huang2017densely} have triggered great interests of applying it to address the environment perception challenges. For instance, deep learning-based computer vision techniques have been increasingly adopted on commercial off-the-shelf advanced driver-assistance systems (ADAS) \cite{tesla,waymo}.

However, recent studies show that deep models (e.g., multilayer perceptrons and convolutional neural networks) are vulnerable to {\em adversarial examples}, which are inputs formed by applying small but crafted perturbations to the {\em clean examples} in order to make the victim deep model yield wrong classification results. This vulnerability is attributed to the linear nature of the deep models \cite{goodfellow6572explaining}. Systematic approaches have been developed to generate adversarial examples as long as the attackers acquire the deep model, where the attackers may know the internals of the model \cite{goodfellow6572explaining} or not \cite{papernot2017practical}. Certain constraints can be considered in the generation process when the attackers cannot tamper with every pixel of the input. For example, in \cite{eykholt2018robust}, an algorithm is developed to determine {\em adversarial stickers} that can be implemented by physically pasting small paper stickers on road signs to mislead vision-based sign classifier. Moreover, as demonstrated in \cite{tencent-2} and explained in \cite{tencent-1}, the vision-based lane detector of Tesla Autopilot, which is an ADAS, can be fooled by small adversarial stickers on the road and thus direct the car to the opposite lane, creating life-threatening danger. Autopilot's weather condition classifier is also shown vulnerable to adversarial examples, leading to wrong operations of the windshield wiper. Therefore, adversarial examples present an immediate and real threat to deep visual sensing systems. The design of these systems must incorporate effective countermeasures especially under the safety-critical settings.

Existing countermeasures aim at hardening the deep models through adversarial training \cite{goodfellow6572explaining,madry2017towards, kannan2018adversarial}, adding a data transformation layer \cite{dziugaite2016study,guo2017countering,das2017keeping,bhagoji2018enhancing,das2018shield,luo2015foveation,xie2017adversarial,wang2016learning,zantedeschi2017efficient}, and gradient masking \cite{song2017pixeldefend,samangouei2018defense,buckman2018thermometer,papernot2016distillation}. These countermeasures are often designed to address certain adversarial examples and build their security on the attackers' ignorance of the defense mechanisms, e.g., the adversarial example generation algorithms used in adversarial training, the data transformation algorithms, and the gradient masking approaches. Thus, they do not address adaptive attackers and fall short of following Kerckhoffs's principle in designing secure systems (i.e., the enemy knows the system except for the secret key \cite{shannon1949communication}). Once the attackers acquire the hardened model and the details of the defense mechanisms, they can craft the next-generation adversarial examples to render the hardened model vulnerable again \cite{athalye2018obfuscated,carlini2019evaluating}.

At present, the deep model training still requires considerable expertise and extensive fine-tuning. As such, in the current practice, a factory-designed deep model is often massively deployed to the products and remains static until the next software upgrade. Such a deployment manner grants the attackers advantage of time. They can extract the deep model (which may be much hardened by the existing countermeasures) from the software release or the memory of a purchased product, study it, and construct the next-generation adversarial examples to affect all products using the same deep model.

Beyond the static defense, in this paper, we consider a {\em moving target defense} (MTD) strategy \cite{jajodia2011moving} to generate one or more new deep models after system deployment that the attackers can hardly acquire.
Different from the identical and static deep model that results in a single point of failure, the generated post-deployment  models are distinct across the systems. This approach invalidates an essential basis for the attackers to construct effective adversarial examples, i.e., the acquisition of the deep model.
Taking the deep visual sensing of ADAS as an example, under the MTD strategy,
new deep models can be continually trained when the computing unit of a car is idle. Once the training completes with the validation accuracy meeting the manufacturer-specified requirement, the new deep models can be commissioned to replace the in-service models that were previously generated on the car.
By bootstrapping the {\em in situ} training with randomness, it will be practically difficult for the attackers to acquire the in-service deep models, which thus can be viewed as the secret of the system. With MTD, the adversarial examples constructed based on a stolen deep model are neither effective across many systems nor effective against a single victim system over a long period of time. In particular, extracting the private deep models from a victim system will require physical access. If the attackers have such physical access, they should launch the more direct and devastating physical attacks that are out of the scope of this paper.

In this paper, we design an MTD approach for embedded deep visual sensing systems that are susceptible to adversarial examples, such as ADAS \cite{boloor2019simple,evtimov2017robust} and biometric authentication \cite{hafemann2019characterizing}. Several challenges need to be addressed. First, adversarial examples have non-negligible transferability to new deep models \cite{szegedy2013intriguing, goodfellow6572explaining, liu2016delving, papernot2016transferability}. From our extensive evaluation based on several datasets, although a new deep model can largely thwart the adversarial examples constructed based on a static {\em base model},
the adversarial examples can still mislead the new deep models with a probability from 7\% to 17\%.
Second, the primitive MTD design of using a single new deep model does not give awareness of the presence of adversarial examples, thus losing the opportunities of involving the human to improve the safety of the system. Note that human can be considered immune to adversarial examples due to their design principle of perturbation minimization. Third, {\em in situ} training of the new deep models without resorting to the cloud is desirable given the concerns of eavesdropping and tampering during the communications over the Internet. However, the training may incur significant computational overhead for the embedded systems.

To collectively address the above challenges, we propose a fork MTD (fMTD) approach based on three key observations on the responses of new deep models to the adversarial examples constructed using the base model. First, the output of a new deep model that is successfully misled by an adversarial example tends to be unpredictable, even though the adversarial example is constructed toward a target label \cite{liu2016delving}. Second, from the minor transferability of adversarial examples and the unpredictability of the misled new model's output, if we use sufficiently many distinct new models to classify an adversarial example, a majority of them will give the correct classification result while the inconsistency of all the models' outputs (due to the unpredictability) signals the presence of attack. This multi-model design echos ensemble machine learning \cite{dietterich2000ensemble}. Third, compared with training a new deep model from scratch, the training with a perturbed version of the base model as the starting point can converge up to 4x faster, imposing less computation burden.

Based on the above observations, we design the fMTD approach as follows. When the system has idle computing resources, it adds independent perturbations to the parameters of the base model to generate multiple {\em fork models}. The base model can be a well factory-designed deep model that gives certified accuracy for clean examples, but may be acquired by the attackers. Each fork model is then used as the starting point of a retraining process. The retrained fork models are then commissioned for the visual sensing task. As the fork models are retrained from the base model, intuitively, they will inherit much of the classification capability of the base model for clean examples. At run time, an input, which may be an adversarial example constructed based on the base model, is fed into each fork model. If the degree of inconsistency among the fork models' outputs exceeds a predefined level, the input is detected as an adversarial example. The majority of the fork models' outputs is yielded as the final result of the sensing task. If the system operates in the human-in-the-loop mode, the human will be requested to classify detected adversarial examples.

The run-time inference overhead of fMTD is proportional to the number of fork models used. Based on our performance profiling on NVIDIA Jetson AGX Xavier, which is a GPU-equipped embedded computing board, instructing TensorFlow to execute the fork models at the same time brings limited benefit in shortening inference time. In contrast, the serial execution of them admits an early stopping mechanism inspired by the serial signal detection \cite{patil2004serial}. Specifically, the system runs the fork models in serial and terminates the execution once sufficient confidence is accumulated to decide the cleanness of the input. Evaluation results show that the serial fMTD reduces the inference time by a factor of up to 5 while achieving the similar sensing and defense performance compared with instructing TensorFlow to execute all fork models.

The contributions of this paper are summarized as follows:
\begin{itemize}
\item Based on important observations on the responses of deep models to adversarial examples, we design fMTD to counteract adversarial example attacks as an ongoing concern.
\item We conduct extensive evaluation on fMTD's accuracy in classifying clean examples as well as its performance in detecting and thwarting adversarial examples under a wide range of settings. The results provide key guidelines for adopters of fMTD in specific applications.
\item We show that the serial execution of the fork models with early stopping can significantly reduce the inference time of fMTD while maintaining the sensing accuracy in both the absence and presence of attacks.
\end{itemize}

The reminder of this paper is organized as follows. \sect\ref{sec:background} reviews background and related work. \sect\ref{sec:measurement} presents a measurement study to motivate the fMTD design. \sect\ref{sec:FMTD} designs fMTD. \sect\ref{sec:evaluation} evaluates the accuracy and attack detection performance of fMTD. \sect\ref{sec:impl} profiles fMTD on Jetson and evaluates the serial fMTD. \sect\ref{sec:discuss} discusses several issues not addressed in this paper. \sect\ref{sec:conclude} concludes this paper.

\section{Background and Related work}
\label{sec:background}

\subsection{Adversarial Examples and Construction}
\label{Adversarial Examples}

Adversarial examples are intentionally designed inputs to mislead deep models to produce incorrect results.
Let $f_{\vecg{\theta}}(\vec{x})$ denote a classifier, where $\vecg{\theta}$ is the classifier's parameter and $\vec{x} \in [0, 1]^m$ is the input (e.g., an image).
Let $y$ denote the ground truth label of $\vec{x}$. The $\vec{x}' = \vec{x} + \vecg{\delta} \in [0, 1]^m$ is an adversarial example, if $f_{\vecg{\theta}}(\vec{x}') \neq y$. The $\vecg{\delta}$ is the perturbation designed by the attackers.
A {\em targeted} adversarial example $\vec{x}'$ makes $f_{\vecg{\theta}}(\vec{x}') = y_t$, where $y_t \neq y$ is a specified target label. A {\em non-targeted} adversarial example ensures that the classification result $f_{\vecg{\theta}}(\vec{x}')$ is an arbitrary label other than the ground truth label $y$.
If the attackers need no knowledge of the classifier's internals (e.g., architecture, hyperparameters, and parameters) to construct the adversarial example, the attack is called {\em black-box} attack. Otherwise, it is called {\em white-box} attack.
In this work, we consider both targeted and non-targeted adversarial examples. As the objective of this paper is to develop a defense approach, it is beneficial to consider the stronger white-box attack, in which the attackers have the knowledge of the internals of the base model.

To increase the stealthiness of the attack to human perception, the difference between $\vec{x}$ and $\vec{x}'$, denoted by $D(\vec{x}, \vec{x}')$, is to be minimized. Thus, the construction of the perturbation for a targeted adversarial example, denoted by $\vecg{\delta}^*_{y_t}$, can be formulated as a constrained optimization problem \cite{szegedy2013intriguing}: $\vecg{\delta}^*_{y_t} = \argmin_{\vecg{\delta}} D(\vec{x}, \vec{x}')$, subject to $f_{\vecg{\theta}}(\vec{x}') = y_t$ and $\vec{x}' \in [0, 1]^m$. The targeted adversarial example that gives the minimum $D(\vec{x}, \vec{x}')$ can be yielded as a non-targeted adversarial example.

Various {\em gradient-based} approaches have been proposed to solve the above constrained optimization problem \cite{szegedy2013intriguing,goodfellow6572explaining,kurakin2016adversarial2, papernot2016limitations, su2019one, carlini2016towards, moosavi2016deepfool, moosavi2017universal, sarkar2017upset, cisse2017houdini, baluja2017adversarial}. Among them, the approach proposed by Carlini and Wagner (C\&W) \cite{carlini2016towards} is often thought to be a highly effective attack construction method and used to evaluate various defense approaches \cite{akhtar2018threat}. We briefly introduce it here. C\&W's approach instantiates the distance to be $\ell_p$-norm and apply Lagrangian relaxation to simplify the problem as: $\vecg{\delta}^*_{y_t} = \argmin_{\vecg{\delta}} \| \vecg{\delta} \|_p + c \cdot g(\vec{x}')$, subject to $\vec{x}' \in [0, 1]^m$, where the regularization function $g(\cdot)$ encompasses the deep model $f_{\vecg{\theta}}(\cdot)$ and satisfies $g(\vec{x}')\geq 0 $ if and only if $f_{\vecg{\theta}}(\vec{x}') = y_t$. The empirical study in \cite{carlini2016towards} shows that the following regularization leads to the most effective attacks in general: $g(\vec{x}') = \max\left\{\max_{y_i \neq y_t} \left\{ Z( \vec{x}')_{y_i} \right\}-Z(\vec{x}')_{y_t},-\kappa \right\}$, where $Z(\cdot)$ represents softmax and $\kappa$ is a parameter controlling the strength of the constructed adversarial example. That is, with a larger $\kappa$, the $\vec{x}'$ is more likely classified as $y_t$. However, the perturbation (or distortion) $\vecg{\delta}$ will be larger. In the inner loop of C\&W's attack construction algorithm for a certain weight $c$, gradient descent is used to solve the constrained optimization problem. Thus, since C\&W's approach exploits the gradients of $f_{\vecg{\theta}}(\cdot)$, it is a gradient-based approach. In the outer loop of the algorithm, binary search is applied to find a setting for $c$ that further minimizes the objective function.
In this paper, we use C\&W's approach to generate adversarial examples and evaluate our fMTD design. Note that the design of fMTD does not rely on any specifics of the C\&W's approach.

\subsection{Countermeasures to Adversarial Examples}

Overfitted models are often thought highly vulnerable to adversarial example attacks. However, regularization approaches for preventing overfitting, such as dropout and weight decay, are shown
ineffective in precluding adversarial examples \cite{szegedy2013intriguing, goodfellow6572explaining}. Brute-force adversarial training \cite{goodfellow6572explaining,madry2017towards, kannan2018adversarial} can improve a deep model to be immune to predefined adversarial examples.
However, it can be defeated by the adversarial examples that are not considered during the adversarial training.
A range of other defense approaches apply various transformations to the input during both the training and inference phases. The transformations include compression \cite{dziugaite2016study, guo2017countering, das2017keeping, bhagoji2018enhancing, das2018shield}, cropping and foveation \cite{luo2015foveation,guo2017countering}, data randomization \cite{xie2017adversarial}, and data augmentation \cite{zantedeschi2017efficient}. These approaches often lead to accuracy drops on clean examples \cite{xu2017feature} and are only effective against the adversarial examples constructed based on the deep model but without the knowledge of the transformation.
{\em Gradient masking} is another category of defense against the adversarial examples constructed using gradient-based methods \cite{song2017pixeldefend,samangouei2018defense,buckman2018thermometer,papernot2016distillation}. It attempts to deny adversary access to useful gradients for constructing attack. However, as shown in \cite{athalye2018obfuscated},
if the attackers know the details of the transformation or the gradient masking, they can still construct effective adversarial examples to fight back. {\em Provable defense} \cite{raghunathan2018certified,dvijotham2018dual,wong2017provable} gives lower bounds of the defense robustness. However, a key limitation of provable defense is that the lower bound is applicable for a set of specific adversarial examples only.

As pointed out by \cite{carlini2019evaluating}, a main drawback of most existing attack prevention and thwarting approaches against adversarial examples is that they do not consider adaptive attackers. Once the attackers acquire the details of the defense, the attackers can bypass the defense. In other words, these approaches' effectiveness is contingent on the attacker's ignorance of the defense mechanism.

In addition to attack prevention and thwarting, adversarial example detection has also received research.
For example, a second classifier can be built to classify an input as clean or adversarial \cite{lu2017safetynet,metzen2017detecting,gong2017adversarial}. Statistical properties of the inputs such principle component have been used to detect attacks \cite{bhagoji2017dimensionality,hendrycks2016early,li2017adversarial}. Others resort to statistical testing \cite{feinman2017detecting,grosse2017statistical}. However, these approaches cannot detect crafty attacks such as the C\&W's attack \cite{carlini2017adversarial}.

\subsection{Moving Target Defense}

Static defense grants attackers the advantage of time. MTD is an emerging approach to address this issue and increase the barrier for effective attacks \cite{jajodia2011moving}.
For instance, the computer hosts can mutate their IP addresses such that the attack traffic is directed to wrong destinations \cite{al2012random}. In the approach described in \cite{sengupta2018mtdeep}, a deep model is randomly selected from a set of candidate models each time to classify an input. The approach uses a limited number of candidate models (e.g., 3 to 6 \cite{sengupta2018mtdeep}) and assumes that they are known to the attackers. Its effectiveness of thwarting the attacks is merely based on the attackers' ignorance of which model is being used, thus following a weak form of MTD. Given the limited number of candidate models, it is not impossible for the attackers to construct an adversarial example that can mislead all candidate models. Moreover, the approach \cite{sengupta2018mtdeep} is short of attack detection capability since a single model is used each time. In contrast, fMTD applies an ensemble of locally generated deep models that can proliferate within the system's available computation resources to achieve both attack detection and thwarting capabilities, thus constituting a strong form of MTD.

\section{Measurement Study}
\label{sec:measurement}

We conduct measurements to gain insights for MTD design.

\subsection{Used Datasets and Deep Models}
\label{subsec:datasets}

We first introduce the datasets and the deep models used in this measurement study as well as the extensive performance evaluation for fMTD in \sect\ref{sec:evaluation}. We use the following three datasets:
\begin{itemize}
  \item {\bf MNIST} \cite{mnist} is a dataset consisting of 60,000 training samples and 10,000 test samples. Each sample is a $28 \times 28$ grayscale image showing a handwritten digit from 0 to 9. We select 5,000 training samples as the validation dataset.
  \item {\bf CIFAR-10} \cite{cifar10} is a 10-class dataset consisting of 50,000 training samples and 10,000 test samples. Each sample is a $32 \times 32$ RGB color image. The 10 classes are airplanes, cars, birds, cats, deers, dogs, frogs, horses, ships, and trucks. We select 5,000 training samples as the validation dataset.
  \item {\bf GTSRB} \cite{Stallkamp-IJCNN-2011} (German Traffic Sign Recognition Benchmark) is a 43-class dataset with more than 50,000 images sizing from $15 \times 15$ to $250 \times 250$ pixels. For convenience, we resize all the images to $32 \times 32$ pixels by interpolation or downsampling. We divide them into training, validation, and test datasets with 34799, 4410, and 12630 samples, respectively.
\end{itemize}

\begin{table}
    \caption{Architectures of the CNNs.}
    \label{tab:model-architecture}
    \begin{tabular}{lll}
    \toprule
     {\bf Layers} & {\bf CNN-A} & {\bf CNN-B} \\ \midrule
     Convolutional + ReLU & $3 \times 3 \times 32$ & $3 \times 3 \times 64$ \\
     Convolutional + ReLU & $3 \times 3 \times 32$ & $3 \times 3 \times 64$ \\
     Max pooling & $2 \times 2 $ & $2 \times 2 $ \\
     Convolutional + ReLU & $3 \times 3 \times 64$ & $3 \times 3 \times 128$ \\
     Convolutional + ReLU & $3 \times 3 \times 64$ & $3 \times 3 \times 128$ \\
     Max pooling & $2 \times 2 $ & $2 \times 2 $ \\
     Fully connected + ReLU & $200$ & $256$ \\
     Fully connected + ReLU & $200$ & $256$ \\
     Softmax & $10$ & $10$ or $43$ \\ \bottomrule
     \multicolumn{3}{l}{$3 \times 3 \times 32$ means 32 $3 \times 3$ convolutional filters.}
    \end{tabular}
    
\end{table}

\begin{table}
  \caption{Training hyperparameters of the CNNs.}
  \label{tab:model-hyper-parameters}
  \begin{tabular}{lll}
  \toprule
   {\bf Hyperparameters} & {\bf CNN-A} & {\bf CNN-B} \\ \midrule
   Learning rate & $0.1$ & $0.01$ (decay $0.5$) \\
   Momentum rate & $0.9$ & $0.9$ (decay $0.5$)\\
   Decay delay & - & $10$ epochs \\
   Dropout rate & $0.5$ & $0.5$ \\
   Batch size & $128$ & $128$ \\
   Epochs & $50$ & $50$ \\ \bottomrule
   \multicolumn{3}{l}{}
  \end{tabular}
\end{table}

We adopt two convolutional neural network (CNN) architectures that have been used in \cite{carlini2016towards} and \cite{papernot2016distillation}. Table~\ref{tab:model-architecture} illustrates the two architectures that are referred to as CNN-A and CNN-B; Table~\ref{tab:model-hyper-parameters} shows the training hyperparameters. 
We apply CNN-A to MNIST. CNN-A has two convolutional layers with 32 $3 \times 3$ filters followed by a max pooling layer, two convolutional layers with 64 $3 \times 3$ filters followed by a max pooling layer, two fully connected layers with 200 rectified linear units (ReLUs) each, and a 10-class softmax layer. CNN-A is trained on MNIST using the momentum-based stochastic gradient descent.
CNN-A achieves training and validation accuracy of 99.84\% and 99.44\%, respectively.

We apply CNN-B to CIFAR-10 and GTSRB. CNN-B's main difference from CNN-A is that more convolutional filters and more ReLUs in the fully connected layers are used to address the more complex patterns of the CIFAR-10 and GTSRB images. Its softmax layer has 10 or 43 classes for CIFAR-10 and GTSRB, respectively.
CNN-B is trained with a learning rate of $0.01$ (decay of 0.5 every 10 epochs) and a momentum rate of $0.9$ (decay of 0.5 every 10 epochs).
For CIFAR-10, CNN-B achieves a validation accuracy of 79.62\%. This result is consistent with those obtained in \cite{carlini2016towards} and \cite{papernot2016distillation}.
For GTSRB, CNN-B achieves training and validation accuracy of 99.93\% and 96.64\%, respectively.

The MNIST and CIFAR-10 datasets have been widely used in image recognition and machine learning research. The use of these two datasets allows us to adopt the CNN architectures that have been shown suitable for them. From the achieved training and validation accuracy, MNIST and CIFAR-10 are representatives of data with simple and complex patterns, respectively. GTSRB gives realism since road sign recognition must be part of ADAS's visual sensing. However, this study does not cater to any specific application. While the detailed results (e.g., classification accuracy) may differ across datasets, we will draw common observations from the results obtained based on these three datasets. The observations will provide useful guidance for the adopters of fMTD to validate their customized fMTD designs in specific applications.

\subsection{Measurement Results}
In this section, we conduct measurements to investigate the responses of multiple {\em new models} to adversarial examples constructed based on the {\em base model} that is different from the new models.

\subsubsection{Adversarial examples}
\label{subsubsec:adversarial-examples}

We use the settings described in \sect\ref{subsec:datasets} to train a base model for each dataset. Then, we use the C\&W approach described in \sect\ref{Adversarial Examples} to generate adversarial examples based on the base model. Specifically, for each dataset, we select a clean test sample in each class as the basis for constructing the targeted adversarial examples whose targeted labels are the remaining classes. For instance, as MNIST has 10 classes, a total of $10 \times 9 = 90$ targeted adversarial examples will be generated. To generate non-targeted adversarial examples for each dataset, we randomly select 100 test samples as the bases for the construction by following the procedure described in \sect\ref{Adversarial Examples}. The C\&W's adversarial examples are highly effective -- all adversarial examples that we generate are effective against the base model.

\begin{table}
  \newlength{\imgw}
  \setlength{\imgw}{0.08\columnwidth}
  \caption{Targeted adversarial examples constructed using C\&W approach \cite{carlini2016towards} with $\ell_2$-norm and various $\kappa$ settings.}
  \label{mnist-change-k}
  \small
  \begin{tabular}{|cc|c|c|c|c|c|c|c|}
    \hline
    & & Clean & \multicolumn{6}{c|}{Attack's target label} \\
    \cline{4-9}
    & & $\!\!\!\!$ example $\!\!\!\!$ & 2 & 3 & 2 & 3 & 2 & 3 \\
    \hline
    \multirow{3}{*}&&&&&&&&\\
    \multirow{3}{*}{\rotatebox[origin=c]{90}{Ground truth label}} & \rotatebox[origin=c]{90}{$\qquad$0} &
    \includegraphics[width=\imgw]{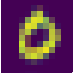} &
    \includegraphics[width=\imgw]{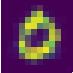} &
    \includegraphics[width=\imgw]{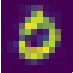} &
    \includegraphics[width=\imgw]{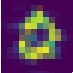} &
    \includegraphics[width=\imgw]{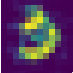} &
    \includegraphics[width=\imgw]{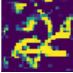} &
    \includegraphics[width=\imgw]{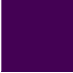} \\
    & \rotatebox[origin=c]{90}{$\qquad$1} & \includegraphics[width=\imgw]{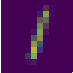} &
    \includegraphics[width=\imgw]{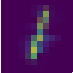} &
    \includegraphics[width=\imgw]{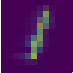} &
    \includegraphics[width=\imgw]{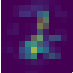} &
    \includegraphics[width=\imgw]{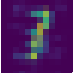} &
    \includegraphics[width=\imgw]{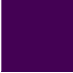} &
    \includegraphics[width=\imgw]{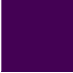} \\
    & \rotatebox[origin=c]{90}{$\qquad$2} & \includegraphics[width=\imgw]{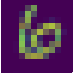} &
    \includegraphics[width=\imgw]{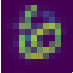} &
    \includegraphics[width=\imgw]{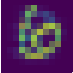} &
    \includegraphics[width=\imgw]{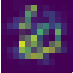} &
    \includegraphics[width=\imgw]{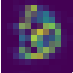} &
    \includegraphics[width=\imgw]{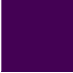} &
    \includegraphics[width=\imgw]{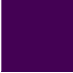} \\
    \hline
    & & & \multicolumn{2}{c|}{$\kappa=0$} & \multicolumn{2}{c|}{$\kappa=45$} & \multicolumn{2}{c|}{$\kappa=95$} \\
    \hline
  \end{tabular}
\end{table}

As described in \sect\ref{Adversarial Examples}, the $\kappa$ is an important parameter of the C\&W's approach that controls the trade-off between the effectiveness of the attack and the distortion introduced. We vary $\kappa$ from 0 to 95. The first image column of Table~\ref{mnist-change-k} shows three clean examples from MNIST. The rest image columns show a number of targeted adversarial examples constructed with three settings of $\kappa$. For instance, all images in the second column will be wrongly classified by the base model as `2'. We can see that with $\kappa=0$, the perturbations introduced by the attack are almost imperceptible to human eyes without referring to the clean examples. With $\kappa=45$, there are clear distortions. With $\kappa=95$, the perturbations may completely erase the figure shapes or create random shapes.

In the rest of this paper, for the sake of attack stealthiness to human, we adopt $\kappa=0$ unless otherwise specified. To confirm the effectiveness of the adversarial examples with $\kappa=0$, we conduct an extended experiment with 1,000 targeted adversarial examples of $\kappa=0$ for MNIST (900 of them are based on $\ell_2$-norm, whereas the remaining are based on $\ell_0$- and $\ell_\infty$-norm due to the slowness in generation). All these 1,000 adversarial examples are effective against the base model.

\subsubsection{Transferability of adversarial examples}
\label{subsubsec:transferability}

\begin{figure}
  \begin{minipage}[t]{.4\columnwidth}
    \vspace{-0.95in}
    \centering
    \makeatletter\def\@captype{table}\makeatother
    \caption{Attack success rate (ASR)}
    \label{tab:attack-success-rate}
    \begin{tabular}{ll}
      \toprule
      MNIST & 6.72\% \\
      CIFAR-10 & 17.3\% \\
      GTSRB & 7.17\% \\
      \bottomrule
      \multicolumn{2}{l}{$^*$ $\kappa=0$.}
    \end{tabular}
  \end{minipage}
  \hfill
  \begin{minipage}[t]{.53\columnwidth}
    \includegraphics{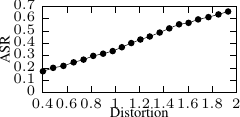}
    \vspace{-1em}
    \caption{ASR vs. distortion.}
    \label{fig:transfer-vs-dist}
  \end{minipage}  
  \vspace{-1em}
\end{figure}

In this set of measurements, for each dataset, we train a new model that has the same architecture as the base model. Then, we measure the attack success rate (ASR) of the adversarial examples on the new model. An adversarial example is successful if the deep model yields a wrong label.
The ASR characterizes the transferability of the adversarial examples to a model differing from the one used for their construction. Table~\ref{tab:attack-success-rate} shows the ASR for the three datasets. We can see that the adversarial examples constructed using the base model can still mislead the new model with probabilities from 7\% to 17\%. This suggests that the adversarial examples have some transferability across different deep models with the same architecture.

We also evaluate the transferability of the adversarial examples constructed with different $\kappa$ settings. We use the Euclidean distance between the adversarial example $\vec{x}'$ and its corresponding clean example $\vec{x}$ to characterize the {\em distortion} caused by the adversarial perturbation. A larger $\kappa$ will result in a larger distortion and thus less stealthiness of the attack to human. Fig.~\ref{fig:transfer-vs-dist} shows the ASR versus distortion for CIFAR-10. We can see that the ASR increases with the distortion. This shows the trade-off between the attack's transferability and stealthiness to human.

\subsubsection{Outputs of multiple new models}
\label{subsubsec:outputs}

\begin{figure}
  \centering
  \begin{minipage}[t]{.48\columnwidth}
    \centering
    \includegraphics{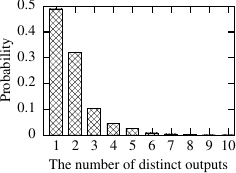}
    \caption{Distribution of the number of distinct outputs of 20 new models given an adversarial example built using the base model.}
    \label{fig:union-size}
  \end{minipage}
  \hfill
  \begin{minipage}[t]{.48\columnwidth}
    \centering
    \includegraphics{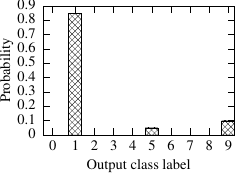}
    \caption{Distribution of the 20 new models' outputs given an adversarial example with ground truth label of 1 and attack target label of 0.}
    \label{fig:prob-vs-class}
  \end{minipage}  
\end{figure}

From \sect\ref{subsubsec:transferability}, adversarial examples have non-negligible transferability to a new model. Thus, using a single new model may not thwart adversarial example attacks. In this set of measurements, we study the outputs of multiple new models. With the base model for each of the three datasets, we construct 270 targeted adversarial examples (i.e., 90 examples based on each of the $\ell_0$, $\ell_2$, and $\ell_\infty$ norms) and 300 non-targeted adversarial examples (i.e., 100 examples based on each of the three norms). For each of the three datasets, we independently train 20 new models.
We denote by $D$ the number of distinct outputs of the 20 models given an adversarial example. Fig.~\ref{fig:union-size} shows the histogram of $D$.
From the figure, the probability that $D$ is greater than one is 51\%. This means that, by simply checking the consistency of the 20 models' outputs, we can detect half of the adversarial example attacks. The probability that $D=1$ (i.e., all 20 new models give the same output) is 49\%. Moreover, 99.5\% of the adversarial examples that result in $D=1$ fail to mislead any new model.
This result suggests that, even if an adversarial example attack cannot be detected by checking the consistency of the 20 models' outputs, it will be thwarted automatically with a high probability.

We now use an example to illustrate whether an adversarial example resulting in $D > 1$ can be thwarted. Fig.~\ref{fig:prob-vs-class} shows the histogram of the 20 new models' outputs given a targeted CIFAR-10 adversarial example with a ground truth label of 1 and a target label of 0. We can see that most new models yield the ground truth label and only a few models yield labels rather than the attack's target label. This shows that the wrong outputs of the new models tend to be unpredictable, rather than the attack's target label. It also suggests that a voting from the distinct outputs of the new models based on a majority rule can thwart the attack.

\subsubsection{Retraining perturbed base model}
\label{subsubsec:retraining}

\begin{table}
  \begin{center}
    \caption{The number of epochs for new model retraining.}
    \label{tab:retrain-epoch}
    \begin{tabular}{|c|c|c|c|}
      \hline
      \multicolumn{1}{|c|}{Intensity of} & \multicolumn{3}{c|}{Dataset} \\
      \cline{2-4}
      \multicolumn{1}{|c|}{perturbation ($w$)} & MNIST & CIFAR-10 & GTSRB \\
      \hline
      0.1 & 11 & 11 & 12 \\
      0.2 & 12 & 13 & 13 \\
      0.3 & 13 & 18 & 13 \\
      training from scratch & 23 & 44 & 22\\
      \hline
    \end{tabular}
  \end{center}
\end{table}

The results in \sect\ref{subsubsec:outputs} suggest that an ensemble of multiple new models is promising for detecting and thwarting adversarial example attacks. However, the training of the new models may incur significant computation overhead. In this section, we investigate a retraining approach. Specifically, we add perturbations to the well trained base model and use the result as the starting point of a retraining process to generate a new model. The model perturbation is as follows.
For each parameter matrix $\vec{M}$ of the base model, we add an independent perturbation to each element in $\vec{M}$. The perturbation is drawn randomly and uniformly from $[w \cdot \min(\vec{M}), w \cdot \max(\vec{M})]$, where $\min(\vec{M})$ and $\max(\vec{M})$ represent the smallest and largest elements of $\vec{M}$, respectively, and $w$ controls the intensity of the perturbation.
The system stops the retraining process if the validation accuracy stops increasing for five consecutive epochs.
Then, the model in the retraining epoch that gives the highest validation accuracy is yielded as a new model.
Table~\ref{tab:retrain-epoch} shows the number of epochs for retraining a new model versus the intensity of the perturbation. We can see that the number of epochs increases with the perturbation intensity. As a comparison, when a new model is trained from scratch with the same stopping criterion, the number of epochs can be up to 4x higher than that with $w=0.1$.

We also measure the time for retraining 20 new models for GTSRB from perturbed versions of the base model on the NVIDIA Jetson computing board. It takes about 45 minutes.

\section{Design of FMTD}
\label{sec:FMTD}

The measurement results in \sect\ref{sec:measurement} suggest an MTD design to counteract adversarial examples. In brief, multiple {\em fork models} can be generated dynamically by retraining independently perturbed versions of the base model. A consistency check on the fork models' outputs can detect whether the input is an adversarial example; the majority of their outputs can be yielded as the final classification result to thwart the adversarial example attack if present.

In this section, we will formally present the system and threat models (\sect\ref{subsec:threat-model}), the design of fMTD that operates autonomously or admits human's input on the detection of an attack (\sect\ref{subsec:fmtd}), and the metrics characterizing the performance of fMTD (\sect\ref{subsec:evaluation-metrics}). The designed fMTD approaches will be extensively evaluated in \sect\ref{sec:evaluation} in terms of these performance metrics.

\subsection{System and Threat Models}
\label{subsec:threat-model}

Consider an embedded visual sensing system (``the system'' for short), which can execute the inference and the training of the used deep model. In this paper, we focus on a single image classification task. Image classification is a basic building block of many visual sensing systems. The classification results can be used to direct the system's actuation. We assume that the system has a well factory-designed model that gives certified accuracy on clean examples and specified adversarial examples. The system also has a training dataset that can be used to train a new deep model locally that achieves a satisfactory classification accuracy as that given by the factory model. Moreover, we make the following two notes regarding the connection of this paper's focus of image classification with the overall visual sensing system in real-world applications.

First, the input to the image classifier may be a cropped area of the original image captured by a camera that contains the object of interest (e.g., the road sign). The cropping can be achieved based on object detection and image segmentation that have received extensive study in computer vision literature \cite{pal1993review}. In this paper, we focus on addressing the adversarial example attacks on the classification task that takes the cropped image as the input. We assume that the object detection and image segmentation work normally. If adversarial example attacks against object detection and image segmentation exist, a separate study will be needed.

Second, some visual sensing systems process a stream of image frames by classifying the individual frames independently and then fusing the classification results over time to yield a final result \cite{han2016seq, tripathi2016context}. In this paper, we focus on the classification of a single image. The attack-proof classification of individual frames will ensure the robustness of the temporal fusion of the classification results. Some other visual sensing systems may take a sequence of image frames as a one-shot input to the deep model \cite{https://calhoun.nps.edu/handle/10945/52976, Broad2018RecurrentMS}. Our MTD approach is also applicable in this setting, since its design does not require specific structure of the deep model's input.

We assume that the attackers cannot corrupt the system. Given that the factory model is static, we assume that the attackers can acquire it via memory extraction, data exfiltration attack, or insiders (e.g., unsatisfied or socially engineered employees). We also assume that the attackers can acquire the training dataset on the system, since the dataset is also a static factory setting. We assume that the attackers can construct stealthy targeted or non-targeted adversarial examples with a white-box approach (e.g., the C\&W approach \cite{carlini2016towards}) based on the factory model or any deep model trained by the attackers using the dataset. Since the focus of this paper is to develop a defense approach, it is beneficial to conservatively consider strong attackers who can launch white-box attacks. This well conforms to Kerckhoffs's principle.

We assume that the system can generate random numbers locally at run time that cannot be acquired by the attackers, although the attackers can know the probabilistic distributions of these random numbers. Truly random number generation can be costly and difficult. Various secure pseudo-random number generation methods can be used instead to achieve practical confidentiality from the attackers. The pseudo-random numbers will be used to perturb the base model and generate fork models. As such, the attackers cannot acquire the exact fork models.

Finally, we assume that the attackers can deploy the adversarial examples, e.g., to paste adversarial paper stickers on road signs.

\subsection{fMTD Work Flow}
\label{subsec:fmtd}

\begin{figure}
  \includegraphics{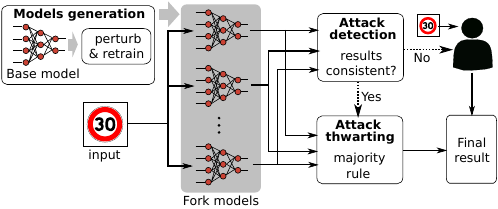}
  \caption{Workflow of fMTD. In the {\em autonomous} mode, the attack thwarting module is executed regardless of the attack detection result. In the {\em human-in-the-loop} mode, the attack thwarting module is executed only when the attack detection gives a positive detection result.}
  \label{fig:FMTD}
\end{figure}

Fig.~\ref{fig:FMTD} overviews the work flow of fMTD. We consider two operating modes of fMTD: {\em autonomous} and {\em human-in-the-loop}. Both modes have the following three components.

\subsubsection{Fork models generation}

To ``move the target'', the system generates new deep models locally for the image classification task. Specifically, we adopt the approach described in \sect\ref{subsubsec:retraining} to perturb the base model with a specified intensity level $w$ and retrain the perturbed model using the training data to generate a fork model.
The retraining completes when the validation accuracy meets a certain criterion.
Using the above procedure, a total of $N$ fork models are generated {\em independently}. We now discuss several issues.

From our evaluation results in \sect\ref{sec:evaluation}, a larger setting of $N$ in general leads to better performance in counteracting the adversarial example attack. Therefore, the largest setting subject to the computation resource constraints and run-time inference timeliness requirements can be adopted. In \sect\ref{sec:impl}, we will investigate the run-time overhead of the fork models.

The fork models generation can be performed right after receiving each new release of the factory-designed model from the system manufacturer. For example, as measured in \sect\ref{subsubsec:retraining}, generating 20 fork models for road sign recognition requires 45 minutes only. To further improve the system's security, the fork models generation can also be performed periodically or continuously whenever the computing unit of the system is idle. For instance, an electric car can perform the generation when it is charging during nights. A newly generated fork model can replace the oldest one among the $N$ fork models.

Since the fork model is retrained from a perturbed version of the base model, the fork model may converge to the base model. However, as the stochastic gradient descent used in the training also incorporates randomness and a deep model often has a large degree of freedom, with a sufficient perturbation intensity level $w$, the fork model is most unlikely identical to the base model.
From a rigorous perspective of information security, the attackers still have a certain amount of information about the fork model since they have the base model and can know the perturbation and retraining mechanisms. Thus, the ensemble of the fork models should be viewed as a quasi-secret of the system only. Nevertheless, MTD is not meant for perfect security, but for significantly increased barriers for the attackers to launch effective attacks.

\subsubsection{Attack detection}

An input is sent to all fork models for classification. From the observations in \sect\ref{subsubsec:outputs}, we can check the consistency of the outputs of all the fork models to detect whether the input is an adversarial example. If more than $T \times 100\%$ of the outputs are the same, the input is detected as a clean example; otherwise, it is detected as an adversarial example. Noted that $T$ is a threshold that can be configured to achieve various satisfactory trade-offs. We will evaluate the impact of $T$ on the performance of the system and discuss its setting in \sect\ref{sec:evaluation}.

\subsubsection{Attack thwarting}

Attack thwarting aims to give the ground truth label of an adversarial example. From the observations in \sect\ref{subsubsec:outputs}, we apply the majority rule to thwart the adversarial example attack. Specifically, the most frequent label among the $N$ fork models' outputs is yielded as the final result.

In the autonomous mode, regardless of the attack detection result, the system will execute the attack thwarting component to generate the final result for the autonomous actuation of the system. Differently, in the human-in-the-loop mode, upon a detection of adversarial example, the system will ask the human operator to classify the input and use the result for the system's subsequent actuation; if no attack is detected, the system will execute the attack thwarting component to yield the final classification result for the subsequent actuation. In this paper, we assume that the human operator will not make any classification error. With this assumption, our performance metrics analysis (\sect\ref{subsec:evaluation-metrics}) and evaluation (\sect\ref{sec:evaluation})  will provide essential understanding on how the human operator's involvement owing to fMTD's attack detection capability improves the system's safety in the presence of attacks. Moreover, since the construction of the adversarial examples follows the perturbation minimization principle to remain imperceptible to human eyes, it is also reasonable to assume that the human operator will not make attack-induced classification error. Nevertheless, our performance metric analysis and evaluation can be easily extended to address human operator's certain error rates when they are non-negligible.

We study both the autonomous and human-in-the-loop modes to understand how the involvement of human affects the system's performance in the absence and presence of adversarial example attacks. Fully autonomous safety-critical systems in complex environments (e.g., self-driving cars) are still grand challenges. For example, all existing off-the-shelf ADAS still requires the driver's supervision throughput the driving process. In this paper, we use the results of the autonomous mode as a baseline.
For either the autonomous or the human-in-the-loop modes, effective countermeasures against adversarial examples must be developed and deployed to achieve trustworthy systems with advancing autonomy.

\subsection{Performance Metrics}
\label{subsec:evaluation-metrics}
\begin{figure}
  \includegraphics[width=\columnwidth]{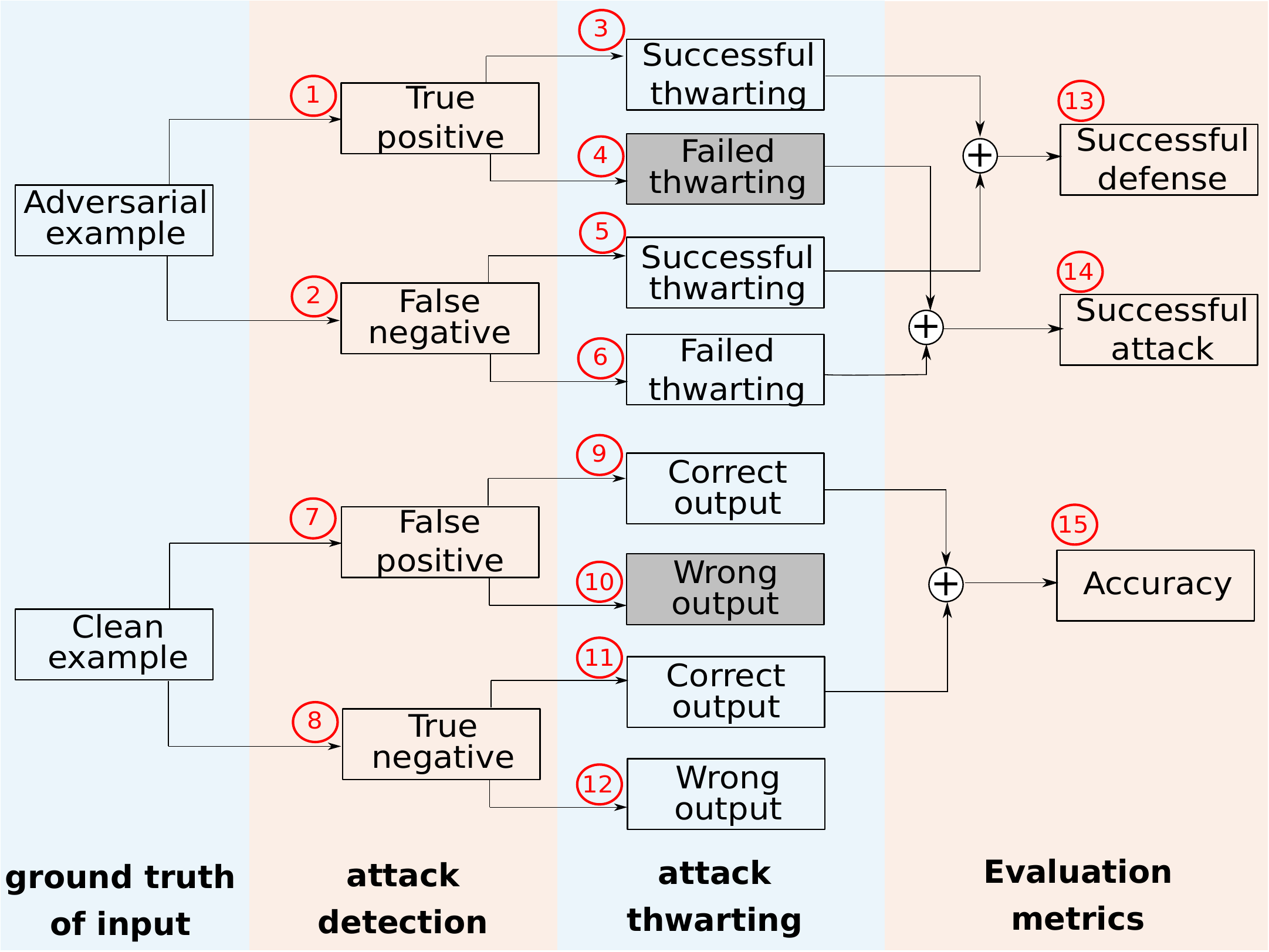}
  
  \caption{Categorization of the system's attack detection and thwarting results and the evaluation metrics. The shaded blocks of ``Failed thwarting'' and ``Wrong output'' are not applicable to human-in-the-loop fMTD.}
  \label{fig:evaluation-metrics}
\end{figure}

In this section, we analyze the metrics for characterizing the performance of fMTD in the autonomous and human-in-the-loop modes. Fig.~\ref{fig:evaluation-metrics} illustrates the categorization of the system's detection and thwarting results. In the following, we use \ciao{$x$} to refer to a block numbered by $x$ in Fig.~\ref{fig:evaluation-metrics}. In \sect\ref{sec:evaluation}, we use $p_x$ to denote the probability of the event described by the block conditioned on the event described by the precedent block. We will illustrate $p_x$ shortly.

When the ground truth of the input is an adversarial example, it may be detected correctly \ciao{1} or missed \ciao{2}. Thus, we use $p_1$ and $p_2$ to denote the true positive and false negative rates in attack detection. We now further discuss the two cases of true positive and false negative:
\begin{itemize}
\item In case of \ciao{1}, the autonomous fMTD may succeed \ciao{3} or fail \ciao{4} in thwarting the attack; differently, the human-in-the loop fMTD can always thwart the attack \ciao{3}. Note that when the attack thwarting is successful, the system will yield the correct classification result; otherwise, the system will yield a wrong classification result.
\item In case of \ciao{2}, the autonomous or human-in-the-loop fMTD may succeed \ciao{5} or fail \ciao{6} in thwarting the attack.
\end{itemize}
The {\em successful defense rate} \ciao{13} is the sum of the probabilities for \ciao{3} and \ciao{5}. The {\em attack success rate} \ciao{14} is the sum of the probabilities for \ciao{4} and \ciao{6}. Note that, with the autonomous fMTD, the two rates are independent of fMTD's detection performance, because the attack thwarting component is always executed regardless of the detection result. In contrast, with the human-in-the-loop fMTD, the two rates depend on fMTD's attack detection performance. In \sect\ref{sec:evaluation}, we will evaluate the impact of the attack detection performance on the two rates.

When the ground truth of the input is a clean example, the detector may generate a false positive \ciao{7} or a true negative \ciao{8}.
\begin{itemize}
\item In case of \ciao{7}, the attack thwarting of the autonomous fMTD may yield a correct \ciao{9} or wrong \ciao{10} classification result; differently, the human-in-the-loop fMTD can always give the correct classification result.
\item In case of \ciao{8}, the attack thwarting of the autonomous or human-in-the-loop fMTD may yield a correct \ciao{11} or wrong \ciao{12} classification result.
\end{itemize}
The {\em accuracy} of the system in the absence of attack \ciao{15} is the sum of the probabilities for \ciao{9} and \ciao{11}.

\begin{figure}
  \newlength{\figw}
  \setlength{\figw}{0.08\columnwidth}
  \small
  \setlength\tabcolsep{1pt}
  \begin{tabular}{cccccccccccc} 
    & & \multicolumn{10}{c}{\bf Attack's target label} \\
    & & 0 & 1 & 2 & 3 & 4 & 5 & 6 & 7 & 8 & 9 \\           
    & \rotatebox[origin=c]{90}{$\qquad$0} &
    \includegraphics[width=\figw]{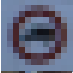} &
    \includegraphics[width=\figw]{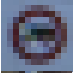} &
    \includegraphics[width=\figw]{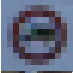} &
    \includegraphics[width=\figw]{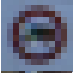} &
    \includegraphics[width=\figw]{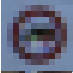} &
    \includegraphics[width=\figw]{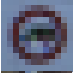} &
    \includegraphics[width=\figw]{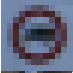} &
    \includegraphics[width=\figw]{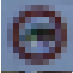} &
    \includegraphics[width=\figw]{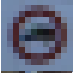} &
    \includegraphics[width=\figw]{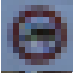} \\[-5pt]
    
    & \rotatebox[origin=c]{90}{$\qquad$1} &
    \includegraphics[width=\figw]{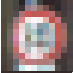} &
    \includegraphics[width=\figw]{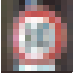} &
    \includegraphics[width=\figw]{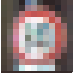} &
    \includegraphics[width=\figw]{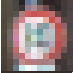} &
    \includegraphics[width=\figw]{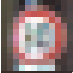} &
    \includegraphics[width=\figw]{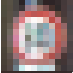} &
    \includegraphics[width=\figw]{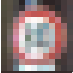} &
    \includegraphics[width=\figw]{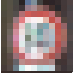} &
    \includegraphics[width=\figw]{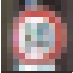} &
    \includegraphics[width=\figw]{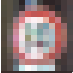} \\[-5pt]

    & \rotatebox[origin=c]{90}{$\qquad$2} &
    \includegraphics[width=\figw]{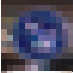} &
    \includegraphics[width=\figw]{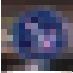} &
    \includegraphics[width=\figw]{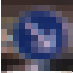} &
    \includegraphics[width=\figw]{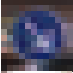} &
    \includegraphics[width=\figw]{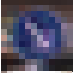} &
    \includegraphics[width=\figw]{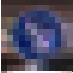} &
    \includegraphics[width=\figw]{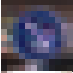} &
    \includegraphics[width=\figw]{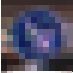} &
    \includegraphics[width=\figw]{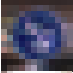} &
    \includegraphics[width=\figw]{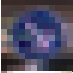} \\[-5pt]

    \multirow{2}{*}{\rotatebox[origin=c]{90}{\bf Ground truth label}}& \rotatebox[origin=c]{90}{$\qquad$3} &
    \includegraphics[width=\figw]{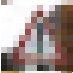} &
    \includegraphics[width=\figw]{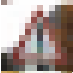} &
    \includegraphics[width=\figw]{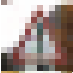} &
    \includegraphics[width=\figw]{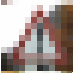} &
    \includegraphics[width=\figw]{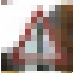} &
    \includegraphics[width=\figw]{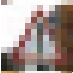} &
    \includegraphics[width=\figw]{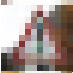} &
    \includegraphics[width=\figw]{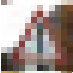} &
    \includegraphics[width=\figw]{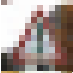} &
    \includegraphics[width=\figw]{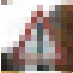} \\[-5pt]

    & \rotatebox[origin=c]{90}{$\qquad$4} &
    \includegraphics[width=\figw]{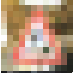} &
    \includegraphics[width=\figw]{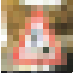} &
    \includegraphics[width=\figw]{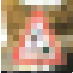} &
    \includegraphics[width=\figw]{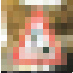} &
    \includegraphics[width=\figw]{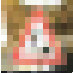} &
    \includegraphics[width=\figw]{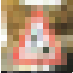} &
    \includegraphics[width=\figw]{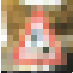} &
    \includegraphics[width=\figw]{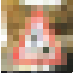} &
    \includegraphics[width=\figw]{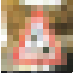} &
    \includegraphics[width=\figw]{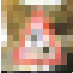} \\[-5pt]

    & \rotatebox[origin=c]{90}{$\qquad$5} &
    \includegraphics[width=\figw]{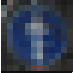} &
    \includegraphics[width=\figw]{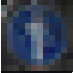} &
    \includegraphics[width=\figw]{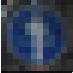} &
    \includegraphics[width=\figw]{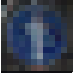} &
    \includegraphics[width=\figw]{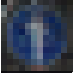} &
    \includegraphics[width=\figw]{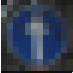} &
    \includegraphics[width=\figw]{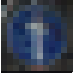} &
    \includegraphics[width=\figw]{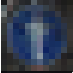} &
    \includegraphics[width=\figw]{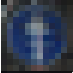} &
    \includegraphics[width=\figw]{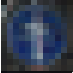} \\[-5pt]
 
    & \rotatebox[origin=c]{90}{$\qquad$6} &
    \includegraphics[width=\figw]{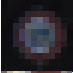} &
    \includegraphics[width=\figw]{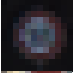} &
    \includegraphics[width=\figw]{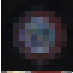} &
    \includegraphics[width=\figw]{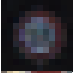} &
    \includegraphics[width=\figw]{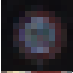} &
    \includegraphics[width=\figw]{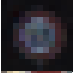} &
    \includegraphics[width=\figw]{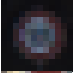} &
    \includegraphics[width=\figw]{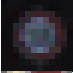} &
    \includegraphics[width=\figw]{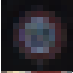} &
    \includegraphics[width=\figw]{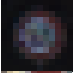} \\[-5pt]

    & \rotatebox[origin=c]{90}{$\qquad$7} &
    \includegraphics[width=\figw]{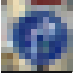} &
    \includegraphics[width=\figw]{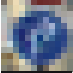} &
    \includegraphics[width=\figw]{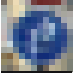} &
    \includegraphics[width=\figw]{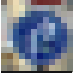} &
    \includegraphics[width=\figw]{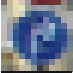} &
    \includegraphics[width=\figw]{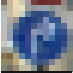} &
    \includegraphics[width=\figw]{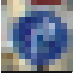} &
    \includegraphics[width=\figw]{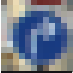} &
    \includegraphics[width=\figw]{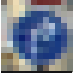} &
    \includegraphics[width=\figw]{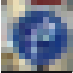} \\[-5pt]

    & \rotatebox[origin=c]{90}{$\qquad$8} &
    \includegraphics[width=\figw]{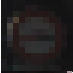} &
    \includegraphics[width=\figw]{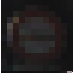} &
    \includegraphics[width=\figw]{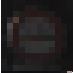} &
    \includegraphics[width=\figw]{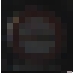} &
    \includegraphics[width=\figw]{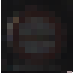} &
    \includegraphics[width=\figw]{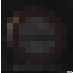} &
    \includegraphics[width=\figw]{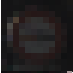} &
    \includegraphics[width=\figw]{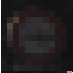} &
    \includegraphics[width=\figw]{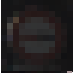} &
    \includegraphics[width=\figw]{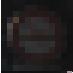} \\[-5pt]

    & \rotatebox[origin=c]{90}{$\qquad$9} &
    \includegraphics[width=\figw]{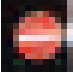} &
    \includegraphics[width=\figw]{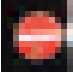} &
    \includegraphics[width=\figw]{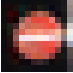} &
    \includegraphics[width=\figw]{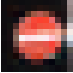} &
    \includegraphics[width=\figw]{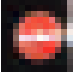} &
    \includegraphics[width=\figw]{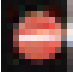} &
    \includegraphics[width=\figw]{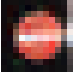} &
    \includegraphics[width=\figw]{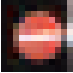} &
    \includegraphics[width=\figw]{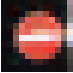} &
    \includegraphics[width=\figw]{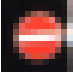} \\[-5pt]

  \end{tabular}
  
  \caption{Targeted adversarial examples constructed using C\&W approach \cite{carlini2016towards} with $\ell_2$-norm and various $\kappa$ settings. Each row consists of adversarial examples generated from the same clean example.}
  \label{fig:adv-targeted-gtsrb-l2}
  
\end{figure}

For fMTD, the successful defense rate $p_{13}$ and the accuracy $p_{15}$ are the main metrics that characterize the system's performance in the presence and absence of attacks. In the autonomous mode, these two metrics are independent of the attack detection performance. Differently, in the human-in-the-loop mode, they are affected by the attack detection performance. In an extreme case, if the detector always gives positive detection results, the human will take over the classification task every time to give the correct results, causing lots of unnecessary burden to the human in the absence of attack. This unnecessary burden can be characterized by the false positive rate $p_7$. There exists a trade-off between this unnecessary burden to human and the system's performance. In summary, the performance of the autonomous fMTD and human-in-the-loop fMTD can be mainly characterized by the tuples of $(p_{13}, p_{15})$ and $(p_7, p_{13}, p_{15})$, respectively.

\section{Performance Evaluation}
\label{sec:evaluation}

In this section, we extensively evaluate fMTD in terms of the performance metrics described in \sect\ref{subsec:evaluation-metrics}.

\subsection{Evaluation Methodology and Settings}

The evaluation is also based on the three datasets described in \sect\ref{subsec:datasets}. We adopt the CNN-A and CNN-B in Table~\ref{tab:model-architecture} and the hyperparameters in Table~\ref{tab:model-hyper-parameters} to train the base models for MNIST, CIFAR-10, and GTSRB. We follow the approach described in \sect\ref{subsubsec:adversarial-examples} to generate the adversarial examples.
Fig.~\ref{fig:adv-targeted-gtsrb-l2} shows a subset of the adversarial examples constructed by the C\&W approach using the $\ell_2$-norm for GTSRB. Note that the images on the diagonal of Fig.~\ref{fig:adv-targeted-gtsrb-l2} are clean examples. We can see that the adversarial perturbations are imperceptible. The fMTD has three configurable parameters: the number of fork models $N$, the model perturbation intensity $w$, and the attack detection threshold $T$. Their default settings are: $N=20$, $w=0.2$, and $T=1$ (i.e., the attack detector will alarm if there is any inconsistency among the fork models' outputs).

\subsection{Results in the Absence of Attack}
\label{subsec:results-absence}

The deployment of the defense should not downgrade the system's sensing accuracy in the absence of attack. This section evaluates this sensing accuracy.
All clean test samples are used to measure the probabilities in the bottom part of Fig.~\ref{fig:evaluation-metrics}.

\begin{figure}
  \begin{subfigure}{.325\columnwidth}
    \includegraphics{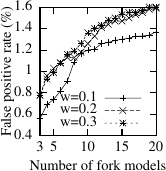}
    \caption{MNIST}
  \end{subfigure}
  \hfill
  \begin{subfigure}{.325\columnwidth}
    \includegraphics{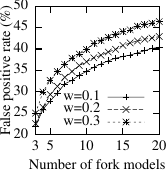}
    \caption{CIFAR-10}
  \end{subfigure}
  \hfill
  \begin{subfigure}{.325\columnwidth}
    \includegraphics{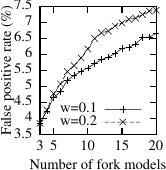}
    \caption{GTSRB}
  \end{subfigure}
  \caption{False positive rate of attack detection ($p_7$).}
  \label{fig:o-7}
\end{figure}

First, we use all the clean test samples to evaluate the false positive rate (i.e., $p_7$) of the attack detection. Fig.~\ref{fig:o-7} shows the measured $p_7$ versus $N$ under various $w$ settings. The $p_7$ increases with $N$. This is because, with more fork models, it will be more likely that the fork models give inconsistent results. Moreover, $p_7$ increases with $w$. This is because, with a higher model perturbation level, the retrained fork models are likely more different and thus give different results to trigger the attack detection. The $p_7$ for CIFAR-10 is more than 20\%. Such a high $p_7$ is caused by the high complexity of the CIFAR-10 images. Moreover, the detector with $T=1$ is very sensitive. With a smaller $T$, the $p_7$ will reduce. For instance, with $T=0.6$, $p_7$ is around 5\%-10\%.

\begin{figure}
  \begin{subfigure}{.325\columnwidth}
    \includegraphics{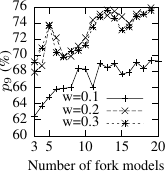}
    \caption{MNIST}
  \end{subfigure}
  \hfill
  \begin{subfigure}{.325\columnwidth}
    \includegraphics{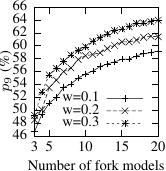}
    \caption{CIFAR-10}
  \end{subfigure}
  \hfill
  \begin{subfigure}{.325\columnwidth}
    \includegraphics{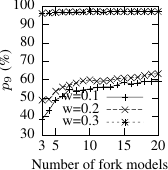}
    \caption{GTSRB}
  \end{subfigure}
  \caption{The rate that the attack thwarting module gives correct output for false positives ($p_9$).}
  \label{fig:o-9}
\end{figure}

\begin{figure}
  \begin{subfigure}{.325\columnwidth}
    \includegraphics{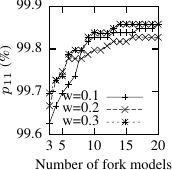}
    \caption{MNIST}
  \end{subfigure}
  \hfill
  \begin{subfigure}{.325\columnwidth}
    \includegraphics{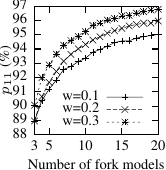}
    \caption{CIFAR-10}
  \end{subfigure}
  \hfill
  \begin{subfigure}{.325\columnwidth}
    \includegraphics{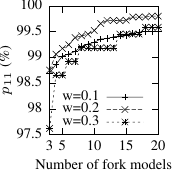}
    \caption{GTSRB}
  \end{subfigure}
  \caption{The rate that the attack thwarting module gives correct output for true negatives ($p_{11}$).}
  \label{fig:o-11}
\end{figure}

Fig.~\ref{fig:o-9} and Fig.~\ref{fig:o-11} show the rates that the attack thwarting gives the correct output when the attack detection yields a false positive and a true negative, respectively. These two rates are $p_9$ and $p_{11}$ as illustrated in Fig.~\ref{fig:evaluation-metrics}. In general, both rates increase with $N$ and $w$. This suggests that with more fork models that are more distinct, the classification performance of the system increases. Another observation is that, $p_9$ is around 60\%, whereas $p_{11}$ is nearly 100\%. This is because, for the clean examples that have triggered the attack detector are not well classifiable; in contrast, the clean examples that have induced all the fork models to produce consistent classification results can be clearly classified.

\begin{figure}
  \begin{subfigure}{.325\columnwidth}
    \includegraphics{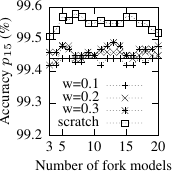}
    \caption{MNIST}
  \end{subfigure}
  \hfill
  \begin{subfigure}{.325\columnwidth}
    \includegraphics{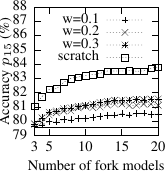}
    \caption{CIFAR-10}
  \end{subfigure}
  \hfill
  \begin{subfigure}{.325\columnwidth}
    \includegraphics{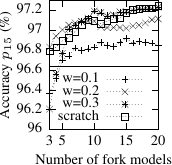}
    \caption{GTSRB}
    \label{fig:accuracy-gtsrb-no-attack}
  \end{subfigure}
  \caption{Accuracy of the system in the absence of attack ($p_{15}$).  The horizontal lines represent the validation accuracy of the respective base models.}
  \label{fig:o-15}
\end{figure}

Fig.~\ref{fig:o-15} shows the accuracy of the system in the absence of attack (i.e., $p_{15}$) versus $N$ under various $w$ settings. The curves labeled ``scratch'' represent the results obtained based on new models trained from scratch, rather than fork models. We can see that training from scratch brings insignificant (less than 2\%) accuracy improvement. The horizontal lines in Fig.~\ref{fig:o-15} represent the validation accuracy of the respective base models. We can see that due to the adoption of multiple deep models, the system's accuracy is improved. This is consistent with the understanding from the decision fusion theory \cite{varshney2012distributed}. The results also show that larger settings for $N$ bring insignificant accuracy improvement. Reasons are as follows.
First, for MNIST and GTSRB, as the accuracy of a single fork model is already high, the decision fusion based on the majority rule cannot improve the accuracy much. Second, for CIFAR-10, although the accuracy of a single fork model is not high (about 80\%), the high correlations among the fork models' outputs impede the effectiveness of decision fusion. Note that, differently, larger settings for $N$ will bring significant defense performance, which will be shown in \sect\ref{subsec:presence-of-attack}.

From Fig.~\ref{fig:accuracy-gtsrb-no-attack}, the accuracy of the road sign recognition is around 97\%. The original images in GTSRB have varied resolutions. To facilitate our evaluation, we resized all the images to $32 \times 32$ pixels. This low resolution contributes to the 3\% error rate. With higher resolutions, this error rate can be further reduced. The main purpose of this evaluation is to show that, in the absence of attacks, fMTD can retain or slightly improve the system's accuracy obtained with the base model. Note that statistical data released by car manufacturers show that ADAS helps reduce safety incident rates \cite{safety-report,waymo-report}, implying the high accuracy of ADAS's visual sensing in the absence of attacks.

\begin{figure}
  \begin{subfigure}{.325\columnwidth}
    \includegraphics{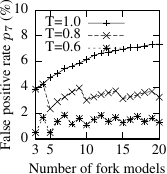}
    \caption{$p_7$ vs. $N$ ($w=0.2$)}
    \label{fig:p7-human-no-attack}
  \end{subfigure}
  \hfill
  \begin{subfigure}{.325\columnwidth}
    \includegraphics{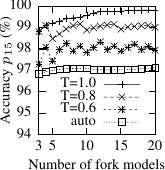}
    \caption{$p_{15}$ vs. $N$ ($w=0.2$)}
    \label{fig:p15-human-no-attack}
  \end{subfigure}
  \begin{subfigure}{.325\columnwidth}
    \includegraphics{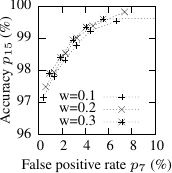}
    \caption{$p_{15}$ vs. $p_7$ ($N=20$)}
    \label{fig:p7-p15-no-attack}
  \end{subfigure}
  \caption{Performance of human-in-the-loop fMTD in the absence of attack. (Dataset: GTSRB)}
  \label{fig:human-no-attack-gtsrb}
  
\end{figure}

Lastly, we consider the human-in-the-loop fMTD. Fig.~\ref{fig:human-no-attack-gtsrb} shows the results based on GTSRB. Specifically, Fig.~\ref{fig:p7-human-no-attack} shows the false positive rate $p_7$ versus $N$ under various settings for the detection threshold $T$. The $p_7$ decreases with $T$, since the attack detector becomes less sensitive with smaller $T$ settings. The $p_7$ characterizes the overhead incurred to the human who will make the manual classification when the attack detector raises an alarm. Fig.~\ref{fig:p15-human-no-attack} shows the accuracy $p_{15}$ versus $N$ under various $T$ settings. The curve labeled ``auto'' is the result for the autonomous fMTD. We can see that the human-in-the-loop fMTD with $T=1$ outperforms the autonomous fMTD by up to 3\% accuracy, bringing the accuracy close to 100\%.
From Fig.~\ref{fig:p7-human-no-attack} and Fig.~\ref{fig:p15-human-no-attack}, we can see a trade-off between the overhead incurred to and the accuracy improvement brought by the human in the loop.
To better illustrate this trade-off, Fig.~\ref{fig:p7-p15-no-attack} shows the accuracy versus the false positive rate under various model perturbation intensity settings. Different points on a curve are the results obtained with different settings of the attack detection threshold $T$. We can clearly see that the accuracy increases with the false positive rate.
In this set of results, the accuracy improvement is from the human's perfect accuracy that we assume. Intuitively, as long as the human's accuracy is higher than the autonomous fMTD, involvement of human is beneficial.

\subsection{Results in the Presence of Attack}
\label{subsec:presence-of-attack}

\begin{figure}
  \begin{subfigure}{.325\columnwidth}
    \includegraphics{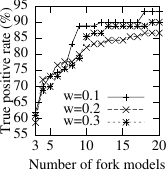}
    \caption{MNIST}
  \end{subfigure}
  \hfill
  \begin{subfigure}{.325\columnwidth}
    \includegraphics{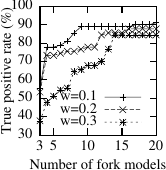}
    \caption{CIFAR-10}
  \end{subfigure}
  \hfill
  \begin{subfigure}{.325\columnwidth}
    \includegraphics{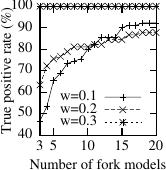}
    \caption{GTSRB}
  \end{subfigure}
  \caption{True positive rate of attack detection ($p_1$).}
  \label{fig:t-1}
\end{figure}

We use the targeted adversarial examples to evaluate the performance of fMTD in detecting and thwarting attacks. Fig.~\ref{fig:t-1} shows the true positive rate (i.e., $p_1$) versus $N$ under various settings of $w$. For the three datasets, the $p_1$ increases from around 50\% to more than 90\% when $N$ increases from 3 to 20. This shows that, due to the minor transferability of adversarial examples, increasing the number of fork models is very effective in improving the attack detection performance. For GTSRB, when $w=0.3$, all attacks can be detected as long as $N$ is greater than 3.

\begin{figure}
  \begin{subfigure}{.325\columnwidth}
    \includegraphics{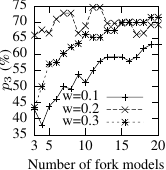}
    \caption{MNIST}
    \label{fig:t-3-mnist}
  \end{subfigure}
  \hfill
  \begin{subfigure}{.325\columnwidth}
    \includegraphics{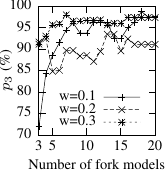}
    \caption{CIFAR-10}
  \end{subfigure}
  \hfill
  \begin{subfigure}{.325\columnwidth}
    \includegraphics{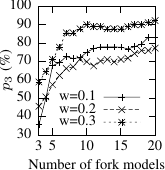}
    \caption{GTSRB}
  \end{subfigure}
  \caption{Rate of successfully thwarting detected attacks ($p_3$).}
  \label{fig:t-3}
\end{figure}

\begin{figure}
  \begin{subfigure}{.325\columnwidth}
    \includegraphics{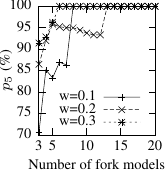}
    \caption{MNIST}
  \end{subfigure}
  \hfill
  \begin{subfigure}{.325\columnwidth}
    \includegraphics{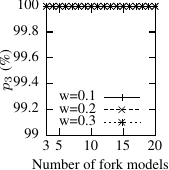}
    \caption{CIFAR-10}
  \end{subfigure}
  \hfill
  \begin{subfigure}{.325\columnwidth}
    \includegraphics{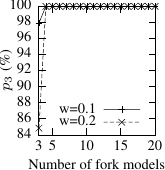}
    \caption{GTSRB}
  \end{subfigure}
  \caption{Rate of successfully thwarting missed attacks ($p_5$).}
  \label{fig:t-5}
\end{figure}

Fig.~\ref{fig:t-3} and Fig.~\ref{fig:t-5} show the rates of successfully thwarting the detected attacks (i.e., $p_3$) and the missed attacks (i.e., $p_5$), respectively. In general, these rates increase with $N$. From the two figures, fMTD is more effective in thwarting the missed attacks than the detected attacks. This is because, for a missed attack, all fork models give the same and correct classification result. However, for the detected attacks, all fork models' results are inconsistent and there is a chance for the majority among the results is a wrong classification result. From Fig.~\ref{fig:t-3-mnist}, MNIST has a relatively low $p_3$. This is because under the same setting of $\kappa=0$, the MNIST adversarial examples have larger distortions. The average distortions introduced by the malicious perturbations, as defined in \sect\ref{subsubsec:adversarial-examples}, are 1.9 and 0.4 for MNIST and CIFAR-10, respectively. Thus, the strengths of the malicious perturbations applied on MNIST are higher, leading to the lower attack thwarting rates in Fig.~\ref{fig:t-3-mnist}.

\begin{figure}
  \begin{subfigure}{.325\columnwidth}
    \includegraphics{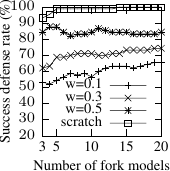}
    \caption{MNIST}
  \end{subfigure}
  \hfill
  \begin{subfigure}{.325\columnwidth}
    \includegraphics{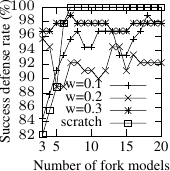}
    \caption{CIFAR-10}
  \end{subfigure}
  \hfill
  \begin{subfigure}{.325\columnwidth}
    \includegraphics{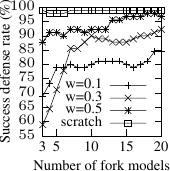}
    \caption{GTSRB}
  \end{subfigure}
  \caption{Successful defense rate ($p_{13}$).}
  \label{fig:t-13}
\end{figure}

Fig.~\ref{fig:t-13} shows the successful defense rate (i.e., $p_{13}$) versus $N$. The $p_{13}$ has an increasing trend with $N$. The curves labeled ``scratch'' represent the results obtained with new models trained from scratch rather than fork models. The fMTD achieves successful defense of 98\% with $w=0.3$ for CIFAR-10 and $w=0.5$ for GTSRB. MNIST has relatively low success defense rates due to the relatively low rates of successfully thwarting detected attacks as shown in Fig.~\ref{fig:t-3-mnist}. However, with new models trained from scratch, the success defense rates for MNIST are nearly 100\%. The higher success defense rates achieved by the new models trained from scratch are due to the lower transferability of adversarial examples to such models. However, training from scratch will incur higher (up to 4x) computation overhead. Thus, there is a trade-off between the attack defense performance and the training computation overhead. We will further discuss this issue in \sect\ref{sec:discuss}.

\begin{figure}
  \begin{subfigure}{.325\columnwidth}
    \includegraphics{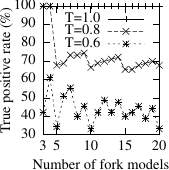}
    \caption{$p_1$ vs. $N$ ($w=0.3$)}
    \label{fig:p1-human-attack}
  \end{subfigure}
  \hfill
  \begin{subfigure}{.325\columnwidth}
    \includegraphics{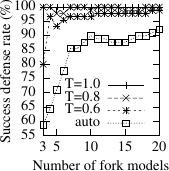}
    \caption{$p_{13}$ vs. $N$ ($w=0.3$)}
    \label{fig:p13-human-attack}
  \end{subfigure}
  \hfill
  \begin{subfigure}{.325\columnwidth}
    \includegraphics{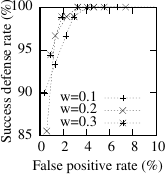}
    \caption{$p_{13}$ vs. $p_7$ ($N=20$)}
    \label{fig:dsr-vs-far-gtsrb}
  \end{subfigure}
  \caption{True positive rate and successful defense rate in the human-in-the-loop mode. (Dataset: GTSRB)}
  \label{fig:human-attack-gtsrb}
\end{figure}

Lastly, we evaluate how the human improves the attack thwarting performance when fMTD operates in the human-in-the-loop mode. Fig.~\ref{fig:human-attack-gtsrb} shows the results based on GTSRB. With a larger $T$ setting (i.e., the detector is more sensitive), the true positive rate increases, requesting more frequent manual classification by the human. As a result, the successful defense rate can increase to 100\%, higher than that of the autonomous fMTD.
Recalling the results in Fig.~\ref{fig:p7-human-no-attack}, a larger $T$ leads to higher false positive rates and thus higher unnecessary overhead incurred to the human. Thus, there exists a trade-off between the successful defense rate and the unnecessary overhead incurred to the human. To better illustrate this trade-off, Fig.~\ref{fig:dsr-vs-far-gtsrb} shows the successful defense rate versus the false positive rate. Different points on a curve are the results obtained with different settings of $T$. We can clearly see that the successful defense rate increases with the false positive rate.

\subsection{Summary and Implication of Results}

From \sect\ref{subsec:results-absence} and \sect\ref{subsec:presence-of-attack}, we have the following observations.

First, in the absence of attack, autonomous fMTD does not improve the classification accuracy much when the number of fork models $N$ increases. Differently, autonomous fMTD's successful defense rate can be substantially improved when $N$ increases. Note that, without fMTD, the adversarial example attacks against the static base model are {\em always} successful. This clearly suggests the necessity of deploying countermeasures.

Second, there exists a trade-off between the successful defense rate and the computation overhead in generating the fork models. Specifically, with more fork models retrained from the base model with larger model perturbation intensity ($w$), higher successful defense rates can be achieved. However, the retraining will have higher computation overhead as shown in Table~\ref{tab:retrain-epoch}. From the results in Fig.~\ref{fig:t-13}, training the new models from scratch gives near-perfect defense performance. However, it incurs computation overhead several times higher than our model forking approach.

Third, the proposed human-in-the-loop design enables the system to leverage the human's immunity to stealthy adversarial examples. The on-demand involvement of human improves the system's accuracy in the absence of attack and the successful defense rate in the presence of attack, with an overhead incurred to the human that is characterized by the false positive rate. From Fig.~\ref{fig:p7-p15-no-attack} and Fig.~\ref{fig:dsr-vs-far-gtsrb} for the GTSRB road sign dataset, with a false positive rate of 4\%, the accuracy without attack is more than 99\% and the successful defense rate is nearly 100\%. The 4\% false positive rate means that, on average, the human will be asked to classify a road sign every 25 clean images of road signs that are detected by ADAS. As adversarial example attacks are rare (but critical) events, how to further reduce the false positive rate while maintaining accuracy and successful defense rate is interesting for further research.

\section{Serial FMTD with Early Stopping}
\label{sec:impl}

In this section, we investigate the run-time overhead of an fMTD implementation on an embedded computing board with hardware acceleration for deep model execution. As many visual sensing systems need to meet real-time requirements, we also investigate how to reduce the run-time overhead of fMTD without compromising its accuracy and defense performance.

\subsection{fMTD Implementation and Profiling}
\label{subsec:impl}

\subsubsection{Setup}

We implement our fMTD approach using TensorFlow and deploy it on an NVIDIA Jetson AGX Xavier \cite{Jetson}. Jetson is an embedded computing board launched in December 2018 and designed for running deep neural networks in applications of automative, manufacturing, retail, and etc.
A Jetson board sizes $10.5 \times 10.5\,\text{cm}^2$ and weighs 280 grams including its thermal transfer plate. It is equipped with an octal-core ARM CPU, a 512-core Volta GPU with 64 Tensor Cores, and 16GB LPDDR4X memory. It runs the Linux4Tegra operating system (version R32.1). The power consumption of Jetson can be configured to be $10\,\text{W}$, $15\,\text{W}$, and $30\,\text{W}$. In our experiments, we configure it to run at $30\,\text{W}$.

\subsubsection{Profiling}

We conduct a set of profiling experiments to compare two possible execution modes of fMTD, i.e., {\em parallel} and {\em serial}. In most deep learning frameworks, the training and testing samples are fed to the deep model in batches. For instance, for ADAS, the road signs segmented from a sequence of frames captured by the camera can form a batch to be fed to the deep model. Our profiling experiments also follow the same batch manner to feed the input samples to the fork models. Specifically, in the parallel mode, a batch of input samples are fed to all fork models simultaneously and all fork models are executed in parallel. This is achieved by the parallel models feature of Keras that is a neural network library running on top of TensorFlow. In the serial mode, a batch of input samples are fed to each of the fork models in serial, i.e., the next model is not executed until the completion of the previous one.

\begin{figure}
    \centering
    \begin{subfigure}[b]{0.48\columnwidth}
        \centering
        \includegraphics[width=\textwidth]{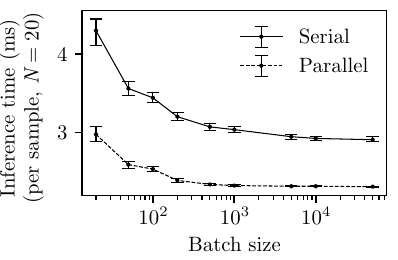}
        \caption{Per-sample inference time}
        \label{fig:time-batch-time}
    \end{subfigure}
    \begin{subfigure}[b]{0.48\columnwidth}
        \centering
        \includegraphics[width=\textwidth]{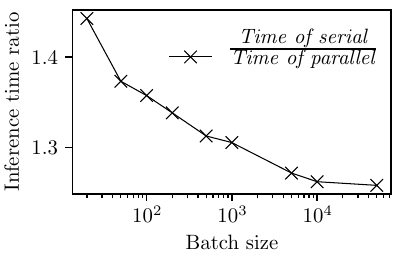}
        \caption{Ratio of inference times}
        \label{fig:time-batch-ratio}
      \end{subfigure}
    \caption{Per-sample inference times of parallel and serial fMTD versus batch size. Error bar represents average, 5th and 95th percentiles over 100 tests under each setting.}
    \label{fig:time-batch}
\end{figure}

We use GTSRB test samples to compare the inference times of the fMTD in the parallel and serial modes.
We vary the settings of the batch size and the number of models. Under each setting, we run fMTD in each mode for 100 times. Fig.~\ref{fig:time-batch-time} shows the per-sample inference time of fMTD with 20 fork models versus the batch size. Each error bar under a batch size setting represents the average, the 5th percentile, and the 95th percentile of the measured inference times.
We can see that the per-sample inference time decreases with the batch size but becomes flat when the batch size is large.
This is because that for a larger batch, TensorFlow can process more samples concurrently. However, with too large batch size settings, the concurrency becomes saturated due to the exhaustion of GPU resources.
The per-sample inference time of the serial fMTD is longer than that of the parallel fMTD. This is because that Keras will try to use all GPU resources to run as many as possible fork models concurrently. Thus, the finite GPU resources will result in a bounded ratio between the inference times of the parallel and serial modes.
Fig.~\ref{fig:time-batch-ratio} shows this ratio versus the batch size. We can see that the ratio decreases from 1.45 to 1.25 and becomes flat. Reason of the decrease is, with larger batch sizes, the parallel mode has less opportunity to really execute the fork models concurrently. This limited ratio suggests that Jetson's GPU resources are still constrained (compared with server platforms). Thus, on resource-constrained embedded platforms with certain hardware acceleration for deep models, the parallel execution of multiple fork models will not reduce the inference time much.

\begin{figure}
    \centering
    \begin{subfigure}[b]{0.48\columnwidth}
        \centering
        \includegraphics[width=\textwidth]{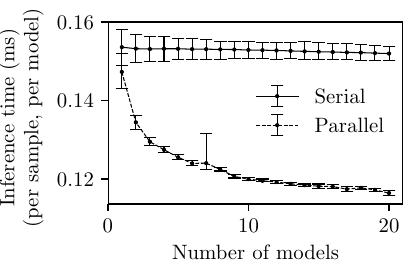}
        \caption{Per-sample per-model inference time vs. $N$}
        \label{fig:time-model-time}
    \end{subfigure}
    \begin{subfigure}[b]{0.48\columnwidth}
        \centering
        \includegraphics[width=\textwidth]{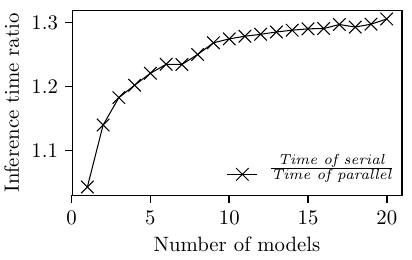}
        \caption{Ratio of inference times vs. the number of models}
        \label{fig:time-model-ratio}
      \end{subfigure}
    \caption{Per-sample per-model inference times of parallel and serial fMTD versus the number of models. Error bar denotes average, 5th and 95th percentiles over 100 tests.}
    \label{fig:time-model}
\end{figure}

Fig.~\ref{fig:time-model-time} shows the per-sample per-model inference time versus the number of fork models $N$. For serial fMTD, the per-sample per-model inference time is independent of $N$. This result is natural. Differently, for parallel fMTD, it decreases with $N$.
Fig.~\ref{fig:time-model-ratio} shows the ratio between the two modes' inference times versus $N$, which increases from about 1 to 1.3. As discussed earlier, because Keras tries to run as many as possible models concurrently, the fMTD with more fork models will be more advantageous in per-model inference time, but will become saturated eventually.

\subsection{Serial fMTD with Early Stopping}

\subsubsection{Design}
From the results in \sect\ref{subsec:impl}, due to the hardware resources constraint, the parallel execution of the fork models does not bring much improvement in terms of inference time. In contrast, the serial execution mode admits early stopping when there is sufficient confidence about the fused result. This is inspired by the serial decision fusion that can reduce the number of decisions needed while maintaining the same event detection performance \cite{patil2004serial}. Algorithm~\ref{alg:fusion} shows the pseudocode of the serial fusion process with early stopping. Note that, in Line~\ref{ln:1}, a subset of three models is the minimum setting enabling the majority-based decision fusion. In Line~\ref{ln:2}, the $T_s$ is a configurable attack detection threshold. We will assess its impact on the serial fMTD's performance shortly. Depending on the operating mode of the system (i.e., autonomous or human-in-the-loop), the classification result of Algorithm~\ref{alg:fusion} is sent to actuation subsystems or the human is requested to perform classification if the input is detected as an adversarial example.

\renewcommand{\algorithmicrequire}{\textbf{Given:}}
\begin{algorithm}
  \caption{Serial fusion with early stopping}
  \label{alg:fusion}
\begin{algorithmic}[1]

\REQUIRE set of fork models $\mathcal{F}$, input $\vec{x}$

\STATE randomly select 3 models from $\mathcal{F}$ and use them to classify $\vec{x}$
\label{ln:1}

\LOOP
\IF{more than $T_s \times 100\%$ of the existing classification results are the same}
\label{ln:2}
\STATE $\vec{x}$ is detected clean and break the loop
\ELSIF{all models in $\mathcal{F}$ have been selected}
\STATE $\vec{x}$ is detected adversarial and break the loop
\ENDIF
\STATE from $\mathcal{F}$ randomly select a model that has not been selected before and use it to classify $\vec{x}$
\ENDLOOP
\RETURN (1) attack detection result and (2) the majority of the existing classification results
\end{algorithmic}
\end{algorithm}

\subsubsection{Evaluation}

\begin{figure}
  \begin{subfigure}{.325\columnwidth}
    \includegraphics{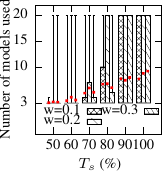}
    \caption{\# of used models}
    \label{fig:used-models}
  \end{subfigure}
  \begin{subfigure}{.325\columnwidth}
    \includegraphics{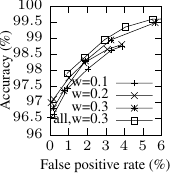}
    \caption{$p_{15}$ vs. $p_7$}
    \label{fig:early-p15-p7}
  \end{subfigure}
  \begin{subfigure}{.325\columnwidth}
    \includegraphics{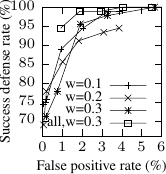}
    \caption{$p_{13}$ vs. $p_7$}
    \label{fig:early-p13-p7}
  \end{subfigure}
  \caption{Performance of human-in-the-loop serial fMTD with early stopping. (Dataset: GTSRB; ``all'' means that early stopping is not enabled; grey line represents median; red square dot represents mean; box represents the (20\%, 80\%) range; upper/lower bar represents maximum/minimum.)}
\end{figure}

In our experiments, we set $N=20$ and vary the serial detection threshold $T_s$ from 0.5 to 1. Fig.~\ref{fig:used-models} shows the number of folk models used in serial fMTD.
For instance, when $T_s \leq 60\%$ and $T_s=100\%$, only three models are used in 95\% and 68\% of all the tests, respectively.
When $T_s = 50\%$ and $T_s=100\%$, 3.08 and 8.37 models are used on average, respectively. The inference times of serial fMTD with $T_s=50\%$ and $T_s=100\%$ are just about 20\% and 50\% of that of parallel fMTD executing all 20 models.

Then, we evaluate the impact of the early stopping on the sensing and defense performance.
Fig.~\ref{fig:early-p15-p7} shows the accuracy ($p_{15}$) versus the false positive rate ($p_7$). Different points on a curve are results under different $T_s$ settings from 0.5 to 1. We can see that, compared with executing all fork models, the early stopping results in little accuracy drop (about 0.1\%).
Fig.~\ref{fig:early-p13-p7} shows the successful defense rate ($p_{13}$) versus the false positive rate ($p_7$). Different points on a curve are results under different $T_s$ settings from 0.5 to 1. We can see that, with a false positive rate of 4\%, the successful defense rate drops 2.2\% only. The above results show that the early stopping can significantly reduce the run-time inference time, with little compromise of accuracy and defense performance. The results for MNIST and CIFAR-10 are similar; we omit them here due to space constraint.

\section{Discussion}
\label{sec:discuss}

The fMTD trains the fork models from perturbed base model. The results in Fig.~\ref{fig:t-13} show that if the new models are trained from scratch, near-perfect defense rates can be achieved. In practice, the factory models can be more sophisticated than the ones used in this paper. The training from scratch may require massive training data and long training time for the embedded system. In addition, the factory models may contain extensive manual tuning by experts. The fMTD's approach of training from perturbed versions of the factory model is more credible to retain the desirable manual tuning. How to retain specific manually tuned features of the factory model in the fork models is interesting to future research.

It is also viable for the manufacturer to generate multiple deep models from the base model for each product independently. If the models remain static after the release of the products, this approach forms a weak form of MTD. The ability of {\em in situ} self-updating of the models is desirable.

\section{Conclusion}
\label{sec:conclude}

This paper presented a fork moving target defense (fMTD) approach for deep learning-based image classification on embedded platforms against the recently discovered adversarial example attacks. We extensively evaluated the performance of fMTD in the absence and presence of attacks. Based on the profiling results of fMTD on NVIDIA Jetson, we also proposed a serial fMTD with early stopping to reduce the inference time. The results in this paper provide useful guidelines for integrating fMTD to the current embedded deep visual sensing systems to improve their security.

\begin{acks}
The authors wish to thank Dr. Yan Xu and Dr. Wentong Cai for their constructive discussions during the development of this work. This research was funded by a Start-up Grant at Nanyang Technological University. We acknowledge the support of NVIDIA Corporation with the donation of two GPUs used in this research.
\end{acks}

\bibliographystyle{ACM-Reference-Format}
\bibliography{001_reference}

\end{document}